\documentclass[sigconf]{aamas} 
\usepackage{balance} % for balancing columns on the final page

\usepackage{breqn}
\usepackage{subfig}
\usepackage{caption}
\usepackage{wrapfig}
\setcopyright{ifaamas}
\acmConference[AAMAS '22]{Proc.\@ of the 21st International Conference
on Autonomous Agents and Multiagent Systems (AAMAS 2022)}{May 9--13, 2022}
{Online}{P.~Faliszewski, V.~Mascardi, C.~Pelachaud,
M.E.~Taylor (eds.)}
\copyrightyear{2022}
\acmYear{2022}
\acmDOI{}
\acmPrice{}
\acmISBN{}

\acmSubmissionID{370}

\title[AAMAS-2022 Formatting Instructions]{Best-Response Bayesian Reinforcement Learning with Bayes-adaptive POMDPs for Centaurs}

\author{Mustafa Mert Çelikok}
\affiliation{
  \institution{Aalto University} \country{Finland}}
\email{mustafamert.celikok@aalto.fi}

\author{Frans A. Oliehoek}
\affiliation{
  \institution{Delft University of Technology} \country{The Netherlands}}
\email{frans.oliehoek@tudelft.nl}

\author{Samuel Kaski}
\affiliation{
  \institution{Aalto University} \country{Finland}}
\affiliation{\institution{University of Manchester} \country{The United Kingdom}}
\email{samuel.kaski@aalto.fi}

\begin{abstract}
Centaurs are half-human, half-AI decision-makers where the AI's goal is to complement the human. To do so, the AI must be able to recognize the goals and constraints of the human and have the means to help them. We present a novel formulation of the interaction between the human and the AI as a sequential game where the agents are modelled using Bayesian best-response models. We show that in this case the AI's problem of helping bounded-rational humans make better decisions reduces to a Bayes-adaptive POMDP. In our simulated experiments, we consider an instantiation of our framework for humans who are subjectively optimistic about the AI's future behaviour. Our results show that when equipped with a model of the human, the AI can infer the human's bounds and nudge them towards better decisions. We discuss ways in which the machine can learn to improve upon its own limitations as well with the help of the human. We identify a novel trade-off for centaurs in partially observable tasks: for the AI's actions to be acceptable to the human, the machine must make sure their beliefs are sufficiently aligned, but aligning beliefs might be costly. We present a preliminary theoretical analysis of this trade-off and its dependence on task structure.
   \end{abstract}


\newcommand{\BibTeX}{\rm B\kern-.05em{\sc i\kern-.025em b}\kern-.08em\TeX}

\begin{document}

%%% The following commands remove the headers in your paper. For final 
%%% papers, these will be inserted during the pagination process.

\pagestyle{fancy}
\fancyhead{}

%%% The next command prints the information defined in the preamble.

\maketitle 

%%%%%%%%%%%%%%%%%%%%%%%%%%%%%%%%%%%%%%%%%%%%%%%%%%%%%%%%%%%%%%%%%%%%%%%%

\section{Introduction}

Humans and AI systems have different computational bounds and biases, and these differences lead to unique strengths and weaknesses. Using this insight, hybrid intelligence aims to combine human and machine
intelligence in a complementary way in order to augment the human intellect \cite{hybridintelligence}. From an agent-based perspective, the hybrid intelligence can be seen as a centaur: a part human, part AI decision-maker. Even though essentially a multiagent team, a distinguishing feature of centaurs is that they appear to others as a single entity, two agents acting as one, when acting in an environment. In this work, we introduce two important design factors for centaurs: I) The AI must be able to learn a model of the human's decision-making, and II) The interaction protocol between the AI and the human must never prohibit an option for the human.

Cognitive science research has been providing empirically verified computational models of human decision-making that can be used as forward and inverse models in control and reinforcement learning settings \cite{ho2021cognitive}. Specifically, the theory of \emph{computational rationality} \cite{https://doi.org/10.1111/tops.12086, gershman2015computational} focuses on developing models of decision-making under computational bounds, resource constraints, and biases. It argues that agents who seem irrational behave rationally according to their subjective models and constraints. This implies that in a centaur, the AI and the human may disagree on optimal behaviour.

\begin{figure}
  \begin{center}
  \includegraphics[width=0.55 \columnwidth]{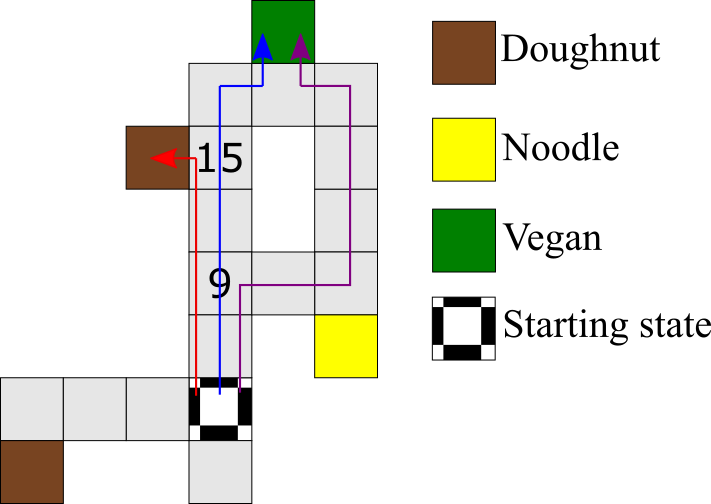}   
  \caption{The Food Truck environment has a vegan restaurant, a noodle café, and two identical doughnut shops. When the preferences are $vegan \succ doughnut \succ noodle$, the blue trajectory is optimal, but the time-inconsistency bias might lead to the red trajectory. In this case, the purple trajectory would help a time-inconsistent human avoid getting tempted by the doughnut shop.}
  \label{fig:single_agent}

  \Description{Image showing the Food Truck environment's grid and possible trajectories.}
 \end{center}
   %\vspace{-3mm}
\end{figure}

Consider the grid-world in Figure \ref{fig:single_agent}, a variant of the \emph{Food Truck} environment of \citet{evans2016learning} where a human's restaurant preferences are $vegan \succ doughnut \succ noodle$. The optimal trajectory is the blue line as the shortest path to the most preferred restaurant. However, the human in question follows the red trajectory: they think they can resist the temptation of a doughnut but when the doughnut shop is too close, they fail to do so, and after the dust settles, they regret their decision. Behavioural sciences explain such behaviour in humans as having preferences that are not consistent over a period of time \cite{Loewenstein1992AnomaliesII, frederick2002time}. In decision-making with delayed rewards, human time-inconsistency is well-modelled by discounting the rewards with hyperbolic functions of the form $d(t;\gamma)=\frac{1}{1+t\gamma}$ \cite{green1994temporal, kirby1995preference, kleinberg2014time}. This is due to the fact that unlike in the exponential discounting of the form $d(t;\lambda) = \lambda^t$, in hyperbolic discounting the ratio $\frac{d(t;\gamma)}{d(t+k;\gamma)}$ depends on $t$ as well as $k$ which can lead to preference reversals.
Now imagine that a time-inconsistent human has recruited an AI to help them eat healthily, and asked the AI to autonomously drive them to the nearest vegan restaurant. Since the typical AI agent has been trained with exponential discounting, it would attempt to follow the blue trajectory, and the human would  override the AI at grid 15 by taking control of the car to stop for a doughnut. However, if the AI was able to predict this, it could try to follow the purple trajectory instead by attempting a detour at grid 9. The purple trajectory costs two time-steps extra, but the human may allow this detour if they decide saving two time-steps is not worth overriding the autonomous driver.

In this paper, we formalize these intuitions by developing a decision-theoretic multiagent model for centaurs. Our main contributions from the most general to specific are: I) We formulate the interaction protocol between the human and the AI as a sequential game. First, the AI proposes to execute an action, and then the human responds by either allowing or overriding it. The human's internal incentive to delegate tasks to the AI is modelled as a cost the human pays when they override an action. We model the agents in a centaur with Bayesian best-response models (defined in Section 3), where learning about the human and nudging them is cast into a Bayesian reinforcement learning problem. (II) We identify a novel challenge in centaurs for partially observable cases, called the \emph{belief alignment} problem, and analyse some of its theoretical properties. (III) As an instance of our framework, we model a class of humans who are optimistic about the machine and compute their best-response approximately. Adopting computational models of human bounded-rationality provided by behavioural economics \cite{Loewenstein1992AnomaliesII, frederick2002time, kleinberg2014time} and cognitive science \cite{green1994temporal,kirby1995preference, 10.3389/fnhum.2012.00100, 10.3389/fpsyg.2019.01676}, we simulate two cases of human bounded-rationality: time-inconsistent preferences and over (or under) estimation of probabilities. Our experiments show that if a human's incentives allow it, the AI is able to nudge the human into making better decisions. The mathematical and computational framework we present opens up a new research direction where various hybrid intelligence problems can be cast into centaurs and more complex model spaces for humans can be investigated in the future.

%%%%%%%%%%%%%%%%%%%%%%%%%%%%%%%%%%%%%%%%%%%%%%%%%%%%%%%%%%%%%%%%%%%%%%%%
\section{Background}
In this section we give a brief introduction to Bayesian reinforcement learning (BRL) for partially observable Markov decision processes (POMDPs), and Bayesian best response models (BA-BRM).

\paragraph{POMDPs and Bayesian Reinforcement Learning.} We define a discrete POMDP by a 6-tuple $(\mathcal{S}, \mathcal{A}, T, R, \Omega, O)$ where $(\mathcal{S}, \mathcal{A}, T, R)$ are the sets of states and actions, transition dynamics, and the reward function of a Markov decision process (MDP); $\Omega$ is a set of observations, and $O: \Omega \times \mathcal{S} \times \mathcal{A} \rightarrow [0,1]$ is the observation function that defines the conditional observation probabilities which can be represented as a set of $|\mathcal{S}| \times |\Omega|$ matrices $\{O^a\}_{\forall a \in \mathcal{A}}$ with the i-th row, j-th column giving the observation probability $O_{(ij)}^a = Pr(o_j \mid s_i, a)$. The Bayes-adaptive POMDP (BA-POMDP) is a BRL model where the unknown transition and observation probabilities are modelled with Dirichlet distributions, and the $\mathcal{S}$ is augmented to include the Dirichlet parameters which transition according to the Bayes rule \cite{Ross2007BayesAdaptiveP}. The solution to the BA-POMDP is the Bayes-optimal policy in terms of the exploration/exploitation trade-off, with respect to the prior.
\paragraph{Bayesian Best Response Models.} In the same vein as interactive POMDPs (I-POMDPs) \cite{GmytrasiewiczPiotr2005AFF} and the recursive modelling method \cite{Gmytrasiewicz2004RationalCI},   Bayesian best-response models (BRM) \cite{Oliehoek14MSDM} fall under the \emph{subjective perspective} approach to multi-agent systems, where the systems are modelled from the point of view of a protagonist agent \cite{Oliehoek2016ACI}. In Bayesian BRMs, the protagonist (denoted by $i$) is considered in its subjective view of the multi-agent environment defined as $MAE_i \triangleq (\mathcal{S}, \{\mathcal{A}_i, \mathcal{A}_j\}, T, \{\Omega_i, \Omega_j \}, O, R_i)$ where $j$ denotes the antagonist, $\mathcal{A}_j$ and $\Omega_j$ are the sets of actions and observations of $j$, and $R_i$ is the protagonist's reward function. The transition dynamics and observation function are defined for the joint set of actions $\mathcal{A} = \mathcal{A}_i \times \mathcal{A}_j$ and the observations $\Omega = \Omega_i \times \Omega_j$. To compute a best-response, the protagonist also needs a model of $j$ as described below, and an \emph{optimality criterion} $OPT_i$ that defines how $i$ accumulates reward over time, such as the exponentially discounted sum of rewards. 

A \emph{model} of an agent $j$ is a tuple $m_j = (\mathcal{A}_j, \Omega_j, \mathcal{I}_j, \pi_j, \beta_j, I_j)$ that is consistent with the $MAE_i$ and fully specifies how the agent will respond to future observations. The $\mathcal{I}_j$ denotes the finite set of internal states of $j$, $I_j$ is its current internal state, $\pi_j: \mathcal{I}_j \rightarrow \Delta(\mathcal{A}_j)$ is a policy mapping the internal states to action probabilities, and $\beta_j: \mathcal{I}_j \times \mathcal{A}_j \times \Omega_j \rightarrow \Delta(\mathcal{I}_j)$ is a belief update function describing how $j$ maintains beliefs over $\mathcal{I}_j$. In a BRM, the agent $i$ is uncertain about some elements of $m_j$ such as $I_j$ or $\pi_j$, however it has a set of models $\mathcal{M}_j$ that are consistent with the known elements of it. In the end, the best-response model combines these three elements, denoted with $BRM_i = (MAE_i, \mathcal{M}_j, OPT_i)$, where the state-space gets augmented to $\Bar{\mathcal{S}} = \mathcal{S} \times \mathcal{I}_j$. The dynamics of the $\textit{BRM}_i$ is defined as:
\begin{align}
    D_i(\Bar{s}'_i, o_i \mid \Bar{s}_i, a_i) = \sum_{a_j} \sum_{o_j} & O(o_j, o_i \mid s', a_i, a_j)T(s' \mid s, a_i, a_j) \nonumber \\ 
    &\beta_j(I'_j \mid I_j, a_j, o_j)\pi_j(a_j \mid I_j).
    \label{eq:brm}
\end{align}
\citet{Oliehoek14MSDM} have shown that BRMs are a class of POMDPs which, when solved, provides $i$'s best response to $j$ . Therefore, if the $T$ and $O$ are unknown to $i$, they can be modelled with Dirichlet distributions as in BA-POMDPs, which leads to Bayes-adaptive best-response models (BA-BRMs). The BA-BRM of $i$ is defined as $\textit{BA-BRM}_i= (\Bar{\mathcal{S}}, \mathcal{A}_i, \Omega_i, \Bar{D}_i, \Bar{R}_i)$ where $\Bar{\mathcal{S}}= \mathcal{S} \times \mathcal{I}_j \times \Theta$ is the augmented set of states with $\Theta$ denoting the parameter space of Dirichlet distributions; $\Bar{R}_i(\Bar{s},a_i)= \sum_{a_j}{R_i(s, a_i, a_j)}\pi_{j}(a_j \mid I_j)$ is the augmented reward function; and $\Bar{D}_i$ is the joint transition and observation dynamics defined as:  
\begin{align}
    &\Bar{D}_{\theta}(\Bar{s}', o_i \mid \Bar{s}, a_i) \triangleq \frac{\theta^{\Bar{s}',o_i}_{\Bar{s},a_i}}{\sum_{\Bar{s}',o_i}{\theta^{\Bar{s}',o_i}_{\Bar{s},a_i}}} \nonumber \\
    &\Bar{D}_i(\Bar{s}', o_i \mid \Bar{s}, a_i) = \Bar{D}_{\theta}(\Bar{s}', o_i \mid \Bar{s}, a_i)\mathbb{I}[\theta' = \theta  + \delta^{\Bar{s}',o_i}_{\Bar{s},a_i}].
    \label{eq:bapomdp}
\end{align}
Here $\theta^{\Bar{s}',o_i}_{\Bar{s},a_i}$ is the count for $(\Bar{s}',o_i,\Bar{s},a_i)$, $\delta^{\Bar{s}',o_i}_{\Bar{s},a_i}$ is the one-hot encoding of $(\Bar{s}',o_i,\Bar{s},a_i)$, and $\mathbb{I}$ is the indicator function. Intuitively, $\textit{BA-BRM}_i$ models the $i$'s uncertainty about $j$ as part of its uncertainty about the dynamics. 

\section{The Human--Machine Centaur (HuMaCe) Model }
In this section we formalize our general framework for human--machine centaurs (HuMaCe), and then present empirical and theoretical results on an instantiation of our framework. The HuMaCe model represents the human and the machine's behaviour as BRMs with different subjective models of the same task, which will be defined explicitly. BRMs inherently assume some model about the other agent, and if that other agent itself employs a BRM this leads to a recursion of beliefs, as in I-POMDPs. Here, we cut this short by considering the BRMs from both agents in isolation, assuming that a sufficiently useful model of the other is available.

\paragraph{The interaction protocol.} The interaction between the human ($h$) and the machine ($m$) is modelled as a sequential game. At time $t$, first the machine chooses an action $a_m$, and the human observes this choice. Then, the human either lets $a_m$ get executed by playing a special \emph{no operation} action $a_h=\textit{noop}$ or overrides it with another action $a_h \neq \textit{noop}$. If the human overrides, they pay an additional cost, $c_h(s, a_h)$, which represents the human's internal incentive to automate the task and delegate things to the machine. When overridden, the machine also receives an additional cost (or reward), $c_m(s, a_h)$, that determines its incentives about getting overridden. The specification of $c_m$ is part of the machine's design and offers significant flexibility. For instance in an autonomous car it makes sense to avoid forcing the human to take control, thus $c_m$ might penalize overrides just like $c_h$. In other cases such as autonomous flight, we may want to incentivize the machine with $c_m$ to safely trigger an override if the human operator is losing attention. The underlying task performed by the centaur is single-agent: either the human or the machine execute an action in the real environment. The executed action, called the \emph{centaur action}, is defined as a function $a_c(a_m, a_h) = \mathbb{I}[a_h \neq \textit{noop}] a_h + \mathbb{I}[a_h = \textit{noop}] a_m$, and it is observed by both agents. This switching-controller property of the protocol leads to MAEs with specific structure where the underlying transition dynamics and rewards are essentially single-agent as $T_i(s' \mid s, a_c(a_m, a_h))$ and $R_i(s, a_c(a_m, a_h))$ for $i \in \{h, m\}$. 

\paragraph{The objective and subjective task models.} As described in the interaction protocol, the underlying dynamics of the agents' multiagent environments is single-agent. Therefore, we represent an agent's sequential decision-making task with a POMDP $M = (\mathcal{S}, \mathcal{A}, T, R, \Omega, O)$ and an \emph{objective optimality criterion} $OPT$. The \emph{objective task model} is defined as $OTM\triangleq(M,OPT)$. When solved exactly, the $OTM$ provides the best behaviour possible (e.g. the blue trajectory in Figure \ref{fig:single_agent}), and represents the ideal problem definition a rational agent can hope for. However, in general the OTM is unknown even to the designer of the machine. Therefore, the human and the machine each have their own subjective model of the task, $\textit{STM}_i$, which consists of their subjective POMDP $\textit{M}_i$ and optimality criterion $\textit{OPT}_i$ for $i \in \{h, m\}$. The $STM$s represent the agents' subjective surrogates that approximate the $OTM$, and since they can be different, they may lead to disagreements on what is optimal behaviour. For instance, in our introductory example the disagreement is due to differences in $\textit{OPT}_i$:  $\textit{OPT}_h$ is the  hyperbolically discounted sum of rewards whereas $OPT_m$ and $OPT$ are exponentially discounted.  

Since the dynamics of MAEs are fully determined by STMs here, an STM and a model of the human suffice to derive the best-response model of the machine.

\subsection{Best Response Model of the Machine}

In this section, we define the machine's model for computing its best response to the human. The model space of the other agent in BRMs is quite general, since it can represent a wide range of agent types, including POMDP-based agents and any policy that can be represented as a finite state controller.

\paragraph{The machine's BRM} We define the model space of the human as $\mathcal{M}_h = \{m_h \mid m_h = (\Bar{\mathcal{A}}_h, \Bar{\Omega}_h, \Bar{O}_h, \mathcal{I}_h, \Bar{\pi}_h, \beta_h, I_h)\}$ where $\Bar{\mathcal{A}}_h = \mathcal{A}_h \cup \{\textit{noop}\}$; the augmented observation space $\Bar{\Omega}_h$ includes $\mathcal{A}_m$, and the $\Bar{O}_h$ provides full observability of the machine's actions. The human's internal states $I_h \in \mathcal{I}_h$ include the observed action of the machine. With the $\mathcal{M}_h$, the machine's BRM is defined as $
    BRM_m(STM_m, \mathcal{M}_h) = ( \Bar{\mathcal{S}}_m, \mathcal{A}_m, \Bar{D}_m, \Bar{\Omega}_m, \Bar{R}_m)$,
where $\Bar{\mathcal{S}}_m = \mathcal{S} \times \mathcal{I}_h$, $\Bar{\Omega}_m = \Omega_m \times \Bar{\mathcal{A}}_h$, and $\Bar{R}_m(\Bar{s}_m, a_m) = R_m(s, a_c(a_m, \Bar{a}_h)) - \mathbb{I}[\Bar{a}_h \neq \textit{noop}]c_m(s, \Bar{a}_h)$. The $\Bar{\Omega}_m$ captures the fact that the machine can observe the human's override, but only after it had happened. The  $\Bar{D}_m$ is defined as follows, where the $a_c$ overloads $a_c = a_c(a_m, \Bar{a}_h)$:
\begin{align}
    \Bar{D}_m(\Bar{s}'_m, \Bar{o}_m \mid \Bar{s}_m, a_m) = \sum_{\Bar{a}_h} \sum_{\Bar{o}_h} & \Bar{O}(\Bar{o}_h, \Bar{o}_m \mid \Bar{s}'_m, a_c)T_m(s' \mid s, a_c)\nonumber \\ 
    &\beta_h(I'_h \mid I_h, a_c, \Bar{o}_h)\Bar{\pi}_h(\Bar{a}_h \mid I_h).
    \label{eq:machinebrm}
\end{align} 
The form of $\Bar{D}_m$ captures the fact that once the machine chooses $a_m$, $\Bar{\pi}_h$ determines the centaur action $a_c$ by overriding or not. Then the dynamics, observations, and the human's belief update evolve according to the $a_c$. The machine cannot observe $\Bar{a}_h$ before it chooses $a_m$, and must predict that using $\Bar{\pi}_h$ and $I_h$. If these two are unknown, then the machine's BRM is augmented to a BA-BRM by placing a prior over them, as described in Section 2.

In practice, the machine will need a more concrete model space for the human in order to infer the human's model and assist them. Next, we define a bounded-rational model space for humans as an instantiation of our framework.

\subsection{The Machine-Optimistic Human}
\label{sec:machine-optimistic-human}
It is not realistic to assume the human will have a perfect model of the machine and will override it optimally with respect to the $STM_h$, due to cognitive bounds. Optimism is one of the most prevalent and ubiquitous approximations humans use to make bounded-rational decisions under cognitive bounds \cite{SHAROT2011R941, Heifetz2000OnTE, automatic_optimism, doi:10.1177/0956797617706706}. We apply this approximation to our setting and propose the \emph{machine-optimistic human (MoH) model}, where the human assumes that the machine's behaviour will agree with them from $t+1$ onward, so after $t$ they will not have to override. This model is one of many possible instantiations of our framework, and future work will focus on enriching the space of human models under the guidance of cognitive sciences as proposed in \citet{ho2021cognitive}.

The following proposition states that a MoH compares the value of the machine's action $a_m$ to the action they think is optimal, using the optimal value function of their $STM_h$: the $Q^{\pi^*_h}$.

\begin{proposition}
Let $Q^{\pi^*_h}(b,a)$ and $V^{\pi^*_h}(b)$ be the value functions of the subjectively optimal policy $\pi^*_h$ computed according to the $OPT_h$ for $M_h$. Given their current belief $b_h$ and the machine's action $a_m$, the MoH overrides the machine if and only if $V^{\pi^*_h}(b_h) - Q^{\pi^*_h}(b_h,a_m) > \mathbb{E}_{s \sim b_h}[c_h(s,\pi^*_{h}(b_h))]$, and if they decide to override $a_m$, the subjectively rational choice is to override with $\pi^*_h(b_h)$.
\end{proposition}
In Supplementary Section 1.1, we prove that Proposition 3.1 follows directly if the human is solving a BRM of their own to decide when to override, using their optimal policy for $STM_h$, the $\pi^*_h$, for predicting how the machine will behave in the future. This result provides a decision-theoretic grounding for the MoH model. 

For the machine, Proposition 3.1 implies that the $Q^{\pi^*_h}$, $b_h$, and $c_h$ are sufficient to predict the MoH's response to $a_m$, where $b_h$ is the MoH's belief in $M_h$ over $\mathcal{S}$. Next, we define the model space for MoH and the machine's BA-BRM for MoH models using this result.

\paragraph{The space of machine-optimistic human models.} A MoH's internal state is defined as $I_h = (b_h, z, a_m)$, where $b_h$ is its belief state in $M_h$ and $a_m$ is the machine's action. The $z$ combines $Q^{\pi^*_h}$ and $c_h$ to a binary function indicating if the MoH will override as: $z(a_m) = \mathbb{I}[V^{\pi^*_h}(b_h) - Q^{\pi^*_h}(b_h,a_m) > \mathbb{E}_{b_h}[c_h(s,\pi^*_{h}(b_h))]]$. The definition of $\Bar{\pi}_h$ follows directly from Proposition 3.1 as $\Bar{\pi}_h(\Bar{a}_h \mid I_h) = z(a_m)\pi^*_h(b_h) + (1 - z(a_m))\delta(\Bar{a}_h, \textit{noop})$ where $\delta$ is the Kronecker delta function. The $\pi^*_h$ is an optimal policy to a POMDP and assumed to be deterministic without loss of generality, and thus $\Bar{\pi}_h$ is also deterministic.

\paragraph{The machine's BA-BRM for machine-optimistic humans.} In our setting the machine's uncertainty over $\mathcal{M}_h$ is due to unknowns $\Bar{\pi}_h$ and $I_h$. However, the human affects the machine's dynamics only through $a_c$ and therefore the machine's uncertainty over $\mathcal{M}_h$ still manifests itself as uncertainty over the dynamics. The most general Bayes-adaptive approach would be to place a Dirichlet prior on the parameters of the categorical distribution $\Bar{D}_m$ as in BA-POMDPs. However, this would ignore the structure in $\Bar{D}_m$: if the machine had known $STM_h = (M_h, OPT_h)$ and the $c_h$, it could compute $Q^{\pi^*_h}$ which is enough to fully determine $\Bar{D}_m$ for MoHs. Therefore, in this case it is more sample-efficient to maintain a posterior over $STM_h$. We will choose an appropriately parameterized distribution $\mu(STM; \theta)$ over the set of possible $STM$s to capture the machine's uncertainty over the MoH's subjective task model. Together with $\mu$, the machine's BA-BRM can be defined as:
\begin{equation}
    \textit{BA-BRM}_m (BRM_m, \mu) = (\Acute{\mathcal{S}}_m, \mathcal{A}_m, \Acute{D}_m, \Bar{\Omega}_m, R_m)
    \label{eq:machinebabrm}
\end{equation}  
where the augmented state space $\Acute{\mathcal{S}}_m=\mathcal{S} \times \mathcal{I}_h \times \Theta$ includes the parameter space of $\mu$ as $\Theta$.

The definition of $\Acute{D}_{m,\theta}(\Acute{s}'_m, \Bar{o}_m \mid \Acute{s}_m, a_m)$ is as follows, where the $\Bar{O}_h$ and $\Bar{\pi}_h$ are fully determined by the corresponding STM:

\begin{align}
     \sum_{\Bar{o}_h} &T_m(s' \mid s, a_c) \Bar{O}_m(\Bar{o}_m \mid \Bar{s}'_m, a_c) \beta_h(I'_h \mid I_h, a_c, \Bar{o}_h) \nonumber \\ 
    &\int\Bar{O}_h(\Bar{o}_h \mid \Bar{s}'_m, a_c, STM) \Bar{\pi}_h(\Bar{a}_h \mid I_h, STM)d\mu(STM;\theta).
    \label{eq:machinebabrmtransition}
\end{align}
The $\Acute{D}_m$ is simply the filtration of $\Acute{D}_{m,\theta}$ similar to the equation \ref{eq:bapomdp}'s $\Bar{D}_i$, which makes sure $\theta$ transitions according to the Bayes rule.

\section{Learning as Planning in HuMaCe}
Any planning algorithm designed for BA-POMDPs can be applied to BA-BRMs, and therefore can solve the machine's problem in HuMaCe \cite{Oliehoek14MSDM}. In this section, we present a specific adaptation of the \emph{root sampling of the model} variant of the BA-POMCP algorithm proposed in \citet{pmlr-v70-katt17a}, for the case of machine-optimistic human models.

\paragraph{An adaptation for machine-optimistic humans} For partially observable state-spaces, each simulation starts by sampling $(s, \theta)$ from the current belief, and in fully-observable settings the current $(s, \theta)$ is known. Then, an $STM$ from $\mu(STM; \theta)$ is sampled, and from thereon the simulation proceeds with the $STM$ fixed. 

Updating our belief about the human's STM $\mu(STM; \theta)$ is intractable, since computing the likelihood requires computing $\Bar{\pi}_h$ for infinitely many $STM$s. However, the (PO)MDPs of STMs are discrete, thus the set of possible $\Bar{\pi}_h$s is finite and discrete for machine-optimistic human models. This implies that for MoH, there are finitely many behavioural equivalence classes over the space of $STM$s. We take advantage of this implication and discretize $\mu$ with a weighted set of particles $\Hat{\mu} = \{STM^{(k)}\}_{k=1,...,N}$. The $\Hat{\mu}$ is initialized by sampling $STM$s according to a prior $\mu(STM; \theta_0)$, and $\Bar{\pi}_h$ is computed and stored offline for each particle. 
\begin{figure}
  \begin{center}
    \includegraphics[width=0.15\textwidth]{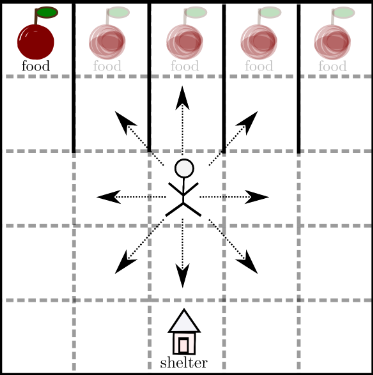}
  \end{center}
  \caption{Food Shelter environment taken from \\ \citet{dimitrakakis2017multi}. Food randomly appears and gives positive reward when consumed. The agent receives a negative reward for each time-step the shelter remains collapsed, and can re-build it by visiting its grid.}
  \label{fig:foodshelter}
  \Description{Image showing the Food Shelter environment.}
\end{figure}The $\mu(STM; \theta)$ therefore is updated by filtering the particles in $\Hat{\mu}$.

\paragraph{Maintaining particles} In some cases, we may need to reinvigorate the particle set $\Hat{\mu}$. New $STM$s can be added by perturbing the existing ones. Unfortunately this will require us to solve the new $STM$s in the online phase. However, if the perturbations applied to existing particles are bounded, it is reasonable to assume the newly added $STM$s will lead to $\Bar{\pi}_h$s that are close to pre-existing ones thanks to behavioural equivalence. In that case, this can be amortized by training a neural network parameterized by $\phi$, $\Hat{\pi}_h(\Bar{a}_h \mid I_h, STM; \phi)$, which takes $(I_h, STM)$ as input and provides a distribution over $\Bar{a}_h$. Further details of this amortization procedure are discussed in Supplementary Section 3.

\section{Experiments and Results}
\label{sec:results}
We consider two cases of bounded-rationality when the humans want the machine to help them make better decisions, and apply our framework instantiated with the machine-optimistic human models. In both cases, the machine's $STM_m$ is a better model of $OTM$ compared to the human's $STM_h$. For the Food Shelter environment, the converse of this case, where the human can correct the machine's behaviour and help it learn a better policy, is also presented. All experiments are run with 19 seeds and both the $c_h(s,a) = c_h$ and $c_m(s,a) = c_m$ are assumed to be non-negative constants. However, the machine does not know $c_h$ and must infer it from interaction. The details of each setting are provided in Supplementary Section 2. 
\subsection{Time-Inconsistent Preferences}

\paragraph{Motivation and setup.} In the introduction, we have given an example for time-inconsistent behaviour: the red trajectory in Figure \ref{fig:single_agent}. We implement the gridworld shown in Figure \ref{fig:single_agent} as an MDP $M$ and the objective optimality criterion $OPT$ is the undiscounted sum of rewards. The subjective task models of agents are $STM_i = (M, OPT_i), i \in \{h,m\}$, where $OPT_h$ is the hyperbolically discounted sum of rewards with discount function $d(t;\gamma)=\frac{1}{1+t\gamma}$, and the $OPT_m$ is exponentially discounted with factor $\lambda=0.95$. Both $\gamma$ and $c_h$ are unknown to the machine. If $\gamma$ is small enough, $d(t;\gamma)$ becomes flat and behaves like exponential discounting, thus this model space is expressive enough to capture both time-inconsistent and consistent behaviour.

In this setting, we do not need the amortization from $\Hat{\pi}_h$. Let $Q^*_\lambda$ denote an optimal Q-function computed with exponential discounting $d(t; \lambda) = \lambda^t$. Any optimal Q-function $Q^*_\gamma$ computed with hyperbolic discounting $d(t;\gamma) = \frac{1}{1+t\gamma}$ is equal to $\int_0^1 w(\gamma, \lambda) Q^*_\lambda d\lambda$ where the weights are computed in closed form as $w(\gamma, \lambda) = \frac{1}{\gamma}\lambda^{\frac{1}{\gamma} - 1}$ \cite{fedus2019hyperbolic}.
We discretize the belief space with a grid of $(c_h, \gamma)$ values, and approximate the integral with a weighted Riemannian sum over a grid of $\lambda$s and their corresponding $Q^*_\lambda$s offline. The details are deferred to Supplementary.

\paragraph{Results.} \begin{figure}
  \begin{center}
    \includegraphics[width=0.33\textwidth]{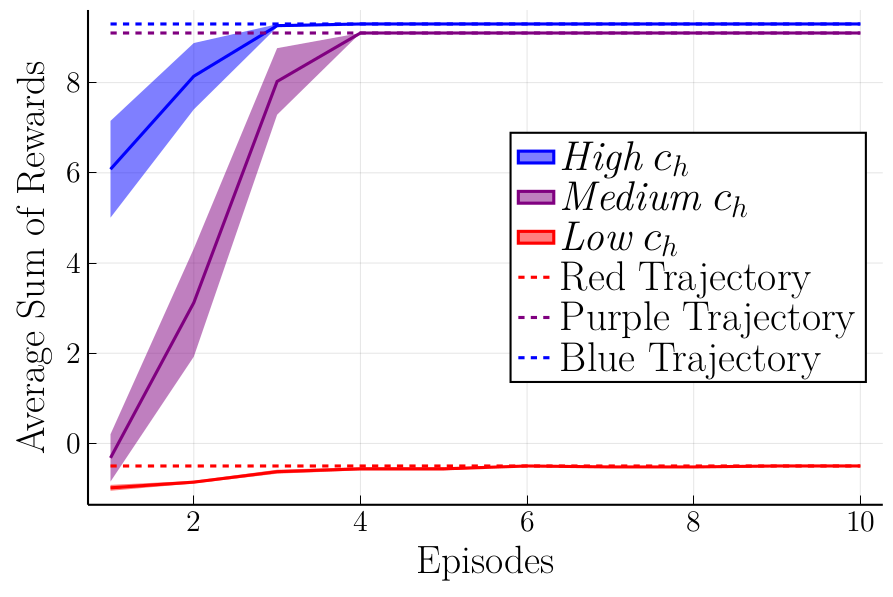}
  \end{center}
  \caption{The machine's return in the Food Truck environment with a time-inconsistent human ($\gamma=7.5$). Low $c_h$ ($0.01$): the human always overrides, so the machine cannot help avoid the red trajectory. Medium $c_h$ ($0.21$): machine follows the purple trajectory and increases the human's performance drastically. Dashed lines show the undiscounted return of the three trajectories from Figure \ref{fig:single_agent} without any overrides. The shades represent one standard error.}
  \label{fig:time_inconsistent_results}
  \Description{Average sum of rewards results for the food truck experiment.}
\end{figure}We simulate the interaction with six behavioural classes of the human based on $(c_h, \gamma)$ pairs: When $\gamma$ is high ($\geq 2.0$), the human's solution to $STM_h$ produces the time-inconsistent red trajectory, and if low ($\leq 0.5$) the hyperbolic function becomes flat and the $OPT_h$ agrees with $OPT_m$, producing the blue trajectory. 
If the $c_h$ is too low ($\leq 0.2$) the AI is overridden whenever the human disagrees, and if high enough ($\geq 0.4$) the AI can execute the blue trajectory regardless of $\gamma$. 
For medium $c_h$ ($[0.21, 0.4]$) and high $\gamma$, the AI cannot execute the blue, but can get the purple accepted. In this case, the Monte Carlo planning in the first episodes attempts to execute the blue trajectory and gets overridden. Once the belief is updated with the override, the following episodes produce the purple trajectory. Figure \ref{fig:time_inconsistent_results} shows the machine's undiscounted return (i.e. $\sum_t{\Bar{r}^{(t)}_{m}}$) for the three cases when $\gamma$ is high. The dashed lines show the returns of trajectories shown in Figure \ref{fig:single_agent}. When $c_h$ is low, the centaur learns that following the red trajectory is the only option, whereas if the $c_h$ is high, it learns to follow the blue. In the case of medium $c_h$, the blue trajectory is not admissible, but the planner learns that the purple is. Thus, the machine improves the human's return drastically. In experiments with low $\gamma$, the centaur follows the blue trajectory since the machine and the human fully agree and the machine never gets overridden.

\subsection{Overestimation of Probabilities} \label{sec:experiment_2}
\paragraph{Motivation and setup.} When using probabilities for making decisions, humans sometimes overestimate the probability of negative events, which may lead to less than ideal decisions, or underestimate, and take risks they cannot afford to. The former may result from a fear and an overestimation of the risk. When overestimation causes the human to unnecessarily avoid certain actions, it is called maladaptive avoidance behaviour \cite{10.3389/fpsyg.2019.01676}.   

We will model the case of overestimation as follows: Given an MDP $M$, the STMs are $STM_i = (M_i, OPT), i \in \{h,m\}$ where the human and the machine disagree in transition probabilities. The $T_h$ is an $\epsilon_{T_h}$-approximation of $T$ with $KL(T(. \mid s,a)\mid\mid T_h(.\mid s,a)) \leq \epsilon_{T_h}$ for all $(s,a)$, whereas, $T_m = T$ which means $STM_m = OTM$. The latter can be relaxed to an $\epsilon_{T_m}$-approximation as long as $\epsilon_{T_h} \geq \epsilon_{T_m}$. A similar setting with a fully-known human model has been investigated by \citet{dimitrakakis2017multi} and we use the same Food Shelter environment from their paper in our experiment, shown in Figure \ref{fig:foodshelter}. Here, food re-appears uniformly randomly and the shelter collapses with non-zero probability. All actions have the same noise where the execution of an action fails with probability $0.1$, transitioning the agent to a neighbour state uniformly random. The human believes that the diagonal action noise is $0.1 + 2\epsilon$ and the rest are $0.1 + \epsilon$. 
\paragraph{Results.} \citet{dimitrakakis2017multi} assume that both the $T_h$ (i.e. $\epsilon$) and $c_h$ are fully known to the machine, but in our experiments they are unknown and must be learned from the interaction. 
\begin{figure}[htp]
  \centering
  \subfloat[The machine's model is correct and the human's is wrong. The \emph{centaur} shows the mean return for our method with unknown $(\epsilon, c_h)$, \emph{naive} is when the machine executes the policy MCTS gives with $STM_m$, ignoring the human. \emph{Ideal} is the machine's MCTS solution to known true $(\epsilon, c_h)$. The shades represent one standard error. \label{fig:foodshelter_results}]{\includegraphics[width=0.35\textwidth]{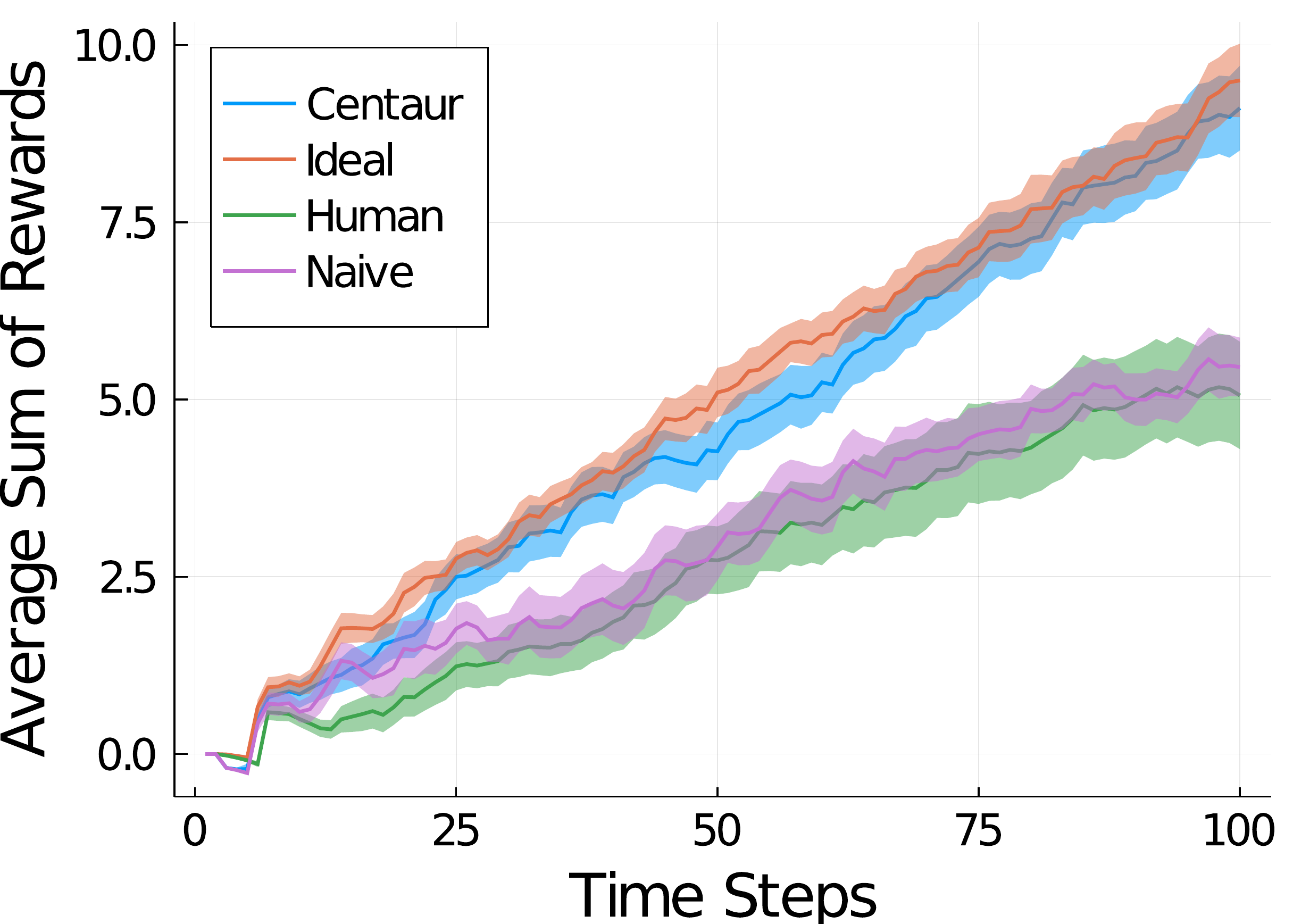} } \hfill
  \subfloat[The machine's model is wrong and the human's is correct. The \emph{human} is the best case policy. The \emph{naive} performs very badly since the machine is overridden many times. \emph{Ideal} knows human's model, and the \emph{centaur} can quickly infer the human's model and avoid getting overridden by learning to perform the better policy. \label{fig:foodshelter_results_machinewrong}]{\includegraphics[width=0.35\textwidth]{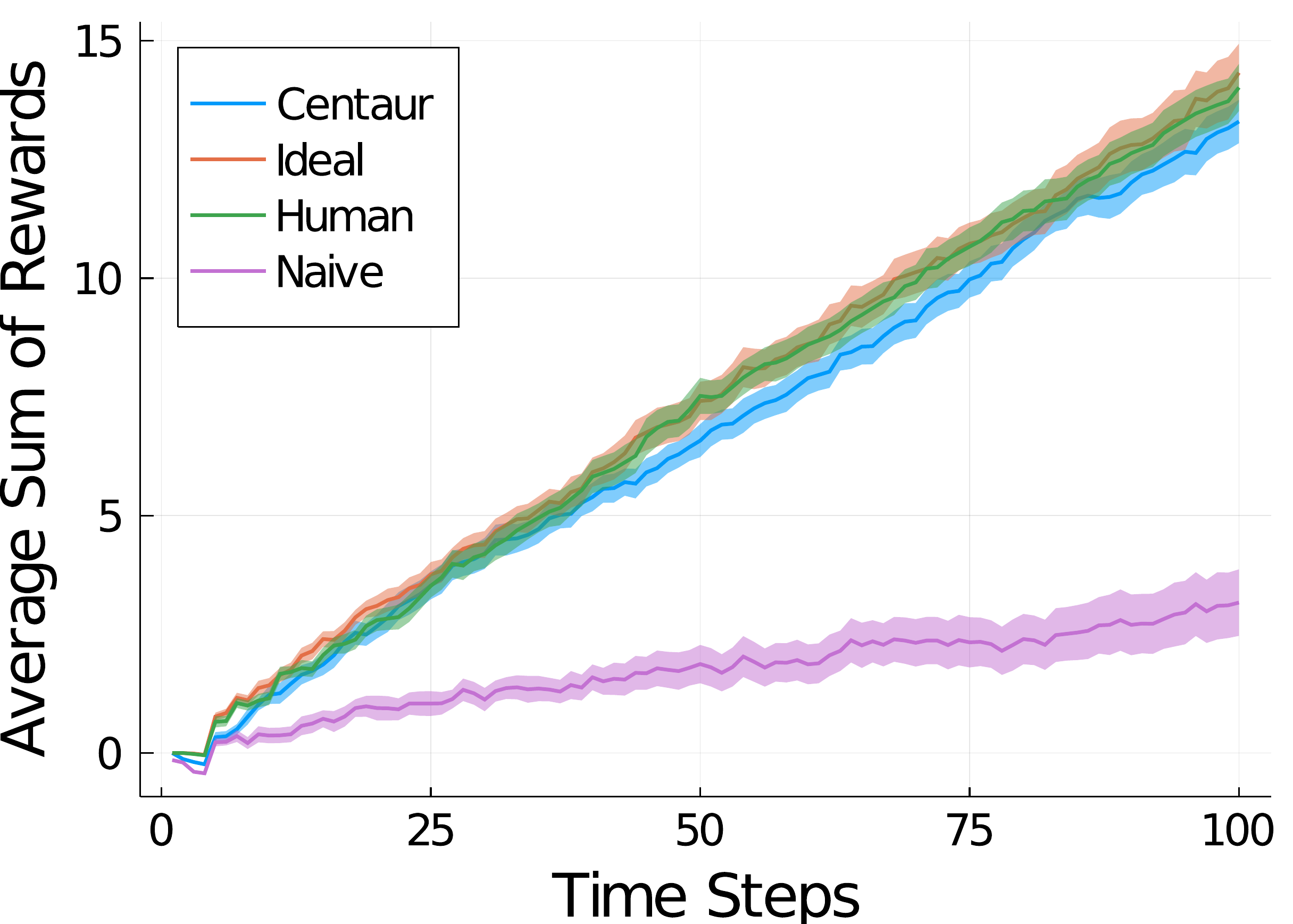}}
  \caption{Machine's return in the Food Shelter environment under overrides of a human, with $(\epsilon=0.45, c_h=0.05, c_m=0.2)$.} \label{fig:foodshelter_general}
  \Description{Empirical results for the food shelter environment.}
  %\vspace{-0.5mm}
\end{figure}

The worst human performance in their experiment is for $\epsilon=0.45$, so we chose this setting and set a low cost of effort $c_h=0.05$. The machine's uniform belief on $\epsilon \in [0.0, 0.45]$ and $c_h \in [0.0, 0.5]$ are discretized covering different behavioural classes. In this case, the initial set of particles had good coverage and re-invigoration was not necessary. We have run the experiment for a fixed horizon of $100$ steps, and noticed that the centaur can infer the true human model $STM_h$ quickly in about the first $10$ time steps. Figure \ref{fig:foodshelter_results} shows the machine's undiscounted return (i.e. $\sum_{t}{\Bar{r}^{(t)}_m}$) evaluated in the real environment (i.e. $OTM$) for three cases: \emph{centaur} is our method, \emph{naive} is when the machine tries to execute its optimal solution to $STM_m$, and \emph{ideal} is when the machine knows $STM_h$ and solves the $BRM_m$ exactly. Additionally, the \emph{human} shows the human's optimal solution to $STM_h$ evaluated in the real environment. The centaur's performance is consistently above the naive and the human, since in certain states it can nudge the human to allow diagonal actions. 

Figure \ref{fig:foodshelter_results_machinewrong} shows the results for the case when the STMs of the two agents are swapped: the human is correct and the machine is wrong. The centaur learns to perform similarly to the human by inferring the human's model and using it to avoid getting overridden. This way, the human effectively teaches the machine to perform a better policy via overrides.

\section{An Analysis of the Partially Observable Cases}
Even though we have defined the HuMaCe and our algorithm in general for partially-observable settings, the $OTM$ and the $STM$s have been fully observable in our empirical results in order to make sure the qualitative aspects of the interaction between the human and the machine are apparent. When both $STM_h$ and $STM_m$ are POMDPs, differences between observation models and transition models can still be converted to differences in transition dynamics in their corresponding fully-observable belief MDPs. However, this does not reduce to the same setting described in Section \ref{sec:experiment_2}. In the fully observable case of Section \ref{sec:experiment_2}, the machine and the human fully agree on the state they are in, but in the case of partial observability the two agents may disagree on this. Specifically, at any given time $t$, the agents' beliefs $(b_{h,t}, b_{m,t})$ may not be the same even if they have the exact same action-observation history. This can happen not only due to differences in transition or observation models, but also due to differences in how the beliefs are maintained. If $b_{h,t} \neq b_{m,t}$, then even when their value functions are the same, the agents may still disagree on what action to take. This is an additional challenge brought by partial observability, and we will call it the \emph{belief alignment problem}.

Here, we present an intuitive example followed by an analysis of the belief alignment problem, and discuss its relation to the task structure. Our analysis reveals a novel trade-off for centaurs: to be able to nudge the human's decisions, the machine must make sure their beliefs are sufficiently aligned. However, aligning beliefs might be costly, and in some cases this trade-off can make it infeasible for the machine to augment the human. This result indicates the importance of studying the partial observability for centaurs in greater detail in future work.

\begin{figure}
  \begin{center}
    \includegraphics[width=0.22\textwidth]{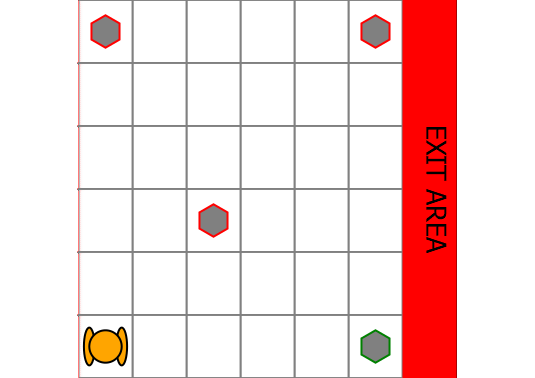}
  \end{center}
  \caption{The RockSample environment. Red rocks are bad quality and greens are good. The rover must pick good rocks and avoid picking bad rocks. The quality of the rocks are partially observed through noisy measurement actions.}
  \label{fig:rocksample}
  \Description{Image showing the RockSample environment.}
\end{figure}

\paragraph{An example for the belief alignment problem.} Consider the RockSample environment in Figure \ref{fig:rocksample}. Here, a rover is rewarded for picking good rock samples and penalized for picking bad ones. The rover knows its position on the map and the position of the rocks, but does not know the quality of the rocks. We denote the fully and partially observed parts of the state variable with $s^k$ (known) and $s^u$ (unknown). The rover can measure any rock from any grid, but the distance to the measured rock increases observation noise exponentially with a rate  determined by the sensor efficiency parameter. When the rover is on top of the rock, the distance is zero, thus there is no noise. Now, let the subjective models of the human and the machine be $STM_h$ and $STM_m$ where the only difference between the two is in the observation models $O_h \neq O_m$: machine thinks the sensor efficiency is quite high whereas the human thinks it is extremely low. Let us also assume that the machine is correct about the true sensor efficiency, therefore $O_m = O$. In this case, since the human thinks observations are extremely noisy, their optimal policy is to visit each rock, measure them on top for a noise-free observation, pick the good ones and then exit. However, the rocks are quite far apart and this wastes time. The machine knows that the true sensor efficiency is much higher, and its optimal policy is to measure all rocks from the starting corner and simply pick up the good ones. Unfortunately, the human does not allow this and overrides. Qualitatively, a similar result to our previous experiments would be if the machine learns a policy that, instead of measuring from the start location, gets closer to the rocks before measuring. This way it can avoid getting overridden, pick up the good rocks, and still save some time compared to the human's policy. Unfortunately, this may not be possible. Since the measurements were not made on top of the rocks, the human considers them very noisy. Therefore, after receiving a "good" observation from a good rock, $b_h$ changes ever slightly while $b_m$ becomes confident that the rock is good, which means the human may still not allow the machine to pick it up. The machine can try to convince the human by measuring the same rock multiple times to get $b_h$ closer to $b_m$, however this can take even longer than simply traversing all the rocks. We note that this issue arises due to the properties of the RockSample environment and not our model. Next, we examine what these properties are, and derive theoretical results intended to guide further research.

\paragraph{Analysis of the RockSample task.} In the rest of this section, we will analyse how the structure of transition and observation dynamics affect the belief alignment problem by using the RockSample as a running example. Let us define two operators $\mathbb{T}^a_i(b)$ and $\mathbb{O}^{o,a}_i(b)$ where $(\mathbb{T}^a_ib)(s) = \sum_{s'}{b(s')T_i(s \mid s',a)}$ and $(\mathbb{O}^{o,a}_ib)(s) = \frac{b(s)O_i(o \mid s)}{\sum_{s'}b(s')O_i(o \mid s')}$. The former propagates a belief through the dynamics model and the latter conditions it on the observation. Therefore, once an action $a$ is taken and observation $o$ is received, rational POMDP agents update their beliefs with $b'= \mathbb{O}^{o,a}_i(\mathbb{T}^a_i(b))$. For the rest of this section, we will assume the human and the machine both have the true dynamics model $T_h = T_m = T$. The dynamics of RockSample is fully deterministic, and the transitions do not change the unknown factor $s^u$ (quality of rocks) at all, which means $\mathbb{T}^a(b) = b$ for all $a$. Moreover, only the measurement actions provide belief changing observations, so $\mathbb{O}^{o,a}_i(b) = b$ for any action other than measurement. The following results specify how the sparsity of belief changing observations and the fact that the transitions do not alter beliefs severely limit the possibility of the two agents aligning their beliefs. 

\begin{definition}[Value of observation \cite{EvenDar2007TheVO}]
Given a hidden Markov model with an observation distribution $O$, let $\mathbf{M}^{O}$ denote the observation matrix with entries $O(o \mid s)$. HMM's value of observation $\gamma(O)$ is defined as $\inf_{x: \mid\mid x \mid\mid_1 =1}\mid\mid \mathbf{M}^Ox \mid\mid_1$. If $\gamma(O) = 0$ the process is unobservable, and if it is $1$, fully observed.
\end{definition}

For a fixed centaur policy $\pi_c$, the subjective POMDPs become hidden Markov models, and the value of observation for each agent depends on the $\pi_c$. In RockSample, if $\pi_c$ does not execute any measurement action, the value of observation is $0$ for both agents and their beliefs do not change. The more measurement actions $\pi_c$ executes, the higher the value of observations gets.

Next, we focus on how the structure of transition dynamics affect agents' belief updates.

\begin{definition}[Minimal mixing rate \cite{Boyen1998TractableIF}]
For a Markovian process with a stochastic transition dynamics $T$, minimal mixing rate $\alpha(T)$ is defined as $\min_{s_1, s_2} \sum_{s'}{\min[T(s' \mid s_1),T(s' \mid s_2)]}$.
\end{definition}

The main result of \citet{Boyen1998TractableIF} is that $KL(\mathbb{T}^a(b_m) \mid \mid \mathbb{T}^a(b_h)) \leq (1 - \alpha(T)) KL(b_m \mid \mid b_h)$. When the transition dynamics are deterministic, the $\alpha(T)=0$, and the contraction result becomes uninformative. In RockSample, it is obvious that since $\mathbb{T}^a(b)=b$, the equality occurs. However, for deterministic dynamics in general, the bound does not tell us whether the contraction happens or not. The following lemma provides a necessary condition for the inequality to be strict in the case of deterministic transitions. Its proof is in Supplementary Section 1.2.

\begin{lemma}
Let $\mathbf{T}^a$ be the deterministic transition matrix with entries $(i,j)$ as $T(s_i \mid s_j, a)$ where $\mathbb{T}^a(b) = \mathbf{T}^ab$. If $rank(\mathbf{T}^a) = \mid S \mid$ for all $a \in \mathcal{A}$, then $KL(\mathbb{T}^a(b_m) \mid \mid \mathbb{T}^a(b_h)) = KL(b_m \mid \mid b_h)$ for all $a \in \mathcal{A}$ and $b_h, b_m \in \Delta(S)$. Therefore, $rank(\mathbf{T}^a) < \mid S \mid$ is a necessary condition for the strict inequality.
\label{lemma:rank}
\end{lemma}

If $rank(\mathbf{T}^a) < \mid S \mid$ for some $a$ but it is $\mid S \mid$ for others, whether the beliefs come closer or not will depend on $\pi_c$. In general, for $\alpha(T)=0$ (deterministic dynamics) and $\gamma(O)=0$ (unobservable), full-rank transition matrices mean beliefs cannot contract. The condition of Lemma \ref{lemma:rank} is indeed satisfied by the RockSample dynamics.  

Next, we combine the value of observation with the minimal mixing rate to derive a bound that describes under what conditions the beliefs of two agents will come closer. The proof of the theorem combines a lemma from \citet{EvenDar2007TheVO} with the theorem of \citet{Boyen1998TractableIF}, and is given in Supplementary Section 1.2.

   \begin{theorem} For  $STM_h = (\mathcal{S}, \mathcal{A},T, \Omega, R, O_h, OPT)$ and $STM_m = (\mathcal{S}, \mathcal{A},T, \Omega, R, O_m, OPT)$ with $O_m = O$, let $KL(O_m(. \mid s,a) \mid \mid O_h(. \mid s,a)) \leq \epsilon_O$ for all $(s,a) \in \mathcal{S} \times \mathcal{A}$. Let  $\mathbf{T}^{a_c}$ be the transition matrix of the hidden Markov model induced by a fixed centaur policy $\pi_c$, with entries $T(s' \mid s, a_c)$ where $a_c$ denotes the executed centaur action. The belief updates satisfy the inequality; 

     \begin{align*}
         \mathbb{E}_{o \sim O_m(. \mid b_m, a_c)}[KL(&b'_m \mid \mid b'_h)] \leq
         \left( 1 - \alpha(\mathbf{T}^{a_c}) \right)KL(b_m \mid\mid b_h) \\ &+\gamma(O_m)3\sqrt{\epsilon_O} -  \left( \frac{\gamma(O_m)KL(b_m \mid \mid b_h)}{\sqrt{2}\log{\frac{1}{\mu}}}\right)^2,
     \end{align*} where $\gamma(O_m)> 0$ is the induced HMM's value of observation, the $\alpha(\mathbf{T}^{a_c})$ is its minimal mixing rate, and $\mu$ is a constant such that $b_h(s), b_m(s) \geq \mu$ for all $s$.
     \label{theorem_1}
   \end{theorem}

Theorem \ref{theorem_1} shows that if the dynamics are deterministic and the $O_m$ is unobservable, the beliefs will not expand. When the $O_m$ is fully observable, the contraction of beliefs depend on how bad the human's approximate observation model is (i.e. $\epsilon_O$). If the minimal mixing rate cannot be influenced (e.g. deterministic dynamics), the only thing machine can do to align beliefs is to increase $\gamma(O_m)$. In RockSample, this means choosing a $\pi_m$ that gets more measurement actions executed. However, the measurement actions cost time, and this trade-off can make it infeasible for the machine to align the human's beliefs with its beliefs sufficiently.

\section{Related Works}

\paragraph{Models and problems related to HuMaCe.} Hybrid intelligence has emerged as a paradigm for designing AI systems that amplify human intelligence and decision-making \cite{hybridintelligence}. Here, we see two main directions of research: (I) Autonomous AI agents that try to influence the collective behaviour of humans positively and (II) Semi-autonomous AI tries to influence its own user via advice, negotiation, or other means. In the former, previous works have shown that self-driving cars can leverage their influence on other human drivers to improve the traffic conditions for every one \cite{Sadigh2016PlanningFA, Kreidieh2018DissipatingSW, Vinitsky2018LagrangianCT, pmlr-v78-wu17a, pmlr-v87-vinitsky18a}.  
This line of work is complementary to ours, which follows the latter direction. Our machine focuses on improving the outcome for its human user by influencing the decisions, similar to behaviour change support systems \cite{OinasKukkonen2012AFF}. 
The closest model to our HuMaCe is the multi-view MDPs and c-intervention games proposed by \citet{dimitrakakis2017multi}, where the human and the machine's MDPs disagree only by transition probabilities and the human may override the machine's action. We generalize this to partially observable environments and other types of disagreements between models. Their interaction protocol is a Stackelberg competition where the machine commits to a policy for the whole horizon, and the human observes the machine's policy and best-responds. In many cases, it is more realistic to assume the human will be able to observe the machine's action, but not know its entire policy, and it may not be possible to make the human act as a follower. Thus, we designed our interaction protocol as a sequential game where the human has the advantage of observing the machine's action and then respond. Their work assumes the full knowledge of the human's $STM_h$ and the cost of effort $c_h$. We generalize this to learning the human's $STM_h$ and $c_h$ from interaction. A similar yet less related model is the Helper Assistant MDP from \citet{fern2014decision}, which only considers the case when the human's goals are unknown and the machine can execute its actions without permission.
\paragraph{Empirical evidence for our models of interactive human behaviour.} \citet{altshuler2018modeling} introduced a sub-optimal decision-making design for when the humans continuously control the autonomy of an assistant system. Their human studies indicate that humans may misjudge the system's performance because their subjective view of the world does not account for high stochasticity. \citet{elmalech2015suboptimal} have demonstrated that in systems that advise humans, sometimes dispersing sub-optimal advice can increase the performance compared to always dispersing optimal advice, because the optimal advice may be too difficult for the human to heed. In the experiments of \citet{10.1145/3359280} for legal decision-making, the machine advice was more accurate than humans' final decisions, and humans were more likely to listen to the advice if they are weakly incentivized. Hyperbolic discounting has been proposed as a good model for time-inconsistent behaviour, consistent with the empirical evidence from human studies \cite{evans2016learning, agentmodels, ainslie2001breakdown, green1994temporal, kirby1995preference}. Previous work in behavioural economics investigated the structure of planning problems where human time-inconsistency may be harmful \cite{kleinberg2014time, 10.1145/2940716.2940764}. The solutions proposed in these works forbid some options of the human by removing them, and therefore not suitable as actions for an AI.
  \balance
\section{Conclusion}
\label{sec:conclusion}
Designers of collaborative AI agents cannot assume that human users will share the same view of the world with the AI. We formalized a general multiagent framework for modelling the decision-making of half-human half-AI agents, \emph{centaurs}, and showed that when equipped with an expressive model space for the human behaviour, the AI can learn how to improve a human's decisions, or improve its own decisions with the help of a human. Cognitive science can provide us with models useful for learning from human decisions. Sufficient statistics can be drawn from these models and used in multiagent reinforcement learning for assisting humans. Finally, we identified a novel trade-off for partially observable cases which highlights the importance of future work on partial observability for centaurs.

\section{Discussions and Future Work}
\label{sec:discussion}

The human in a centaur is modelled as a decision-maker who \emph{may} be imperfect due to their bounded-rationality. Of course, all models of humans are wrong, but some are useful. We need not claim that our method can identify the \emph{true model} of human decisions, just that given a good model space it can infer a useful model for assistance. In our interaction protocol, the AI cannot prohibit any trajectory. For instance, completely ignoring the AI and committing to always overriding is an admissible strategy for the human in the sequential game. We consider the human as an independent, self-interested agent, and the $c_h$ comes from their own incentives. This formulation can capture various human factors in automation, such as automation misuse where the human is over-reliant on the AI (very high $c_h$) or automation disuse (very low $c_h$) \cite{parasuraman1997humans}. The STMs allow us to capture a rich set of differences between humans and AI systems. Since the STMs are surrogates for an objective task model, the subjective views of agents are not arbitrarily different. This is both realistic and computationally useful, since it reduces the size of the model space of humans.

\paragraph{Limitations and future work.} BRMs can capture a wide range of behaviours, including bounded-rationality, and our experiments indicate that even infinite model spaces may lead to finite behavioural equivalence classes. Our framework is not limited to machine-optimistic humans, and future work can consider more advanced models for the human such as learning agents with bounded adaptation capabilities to the machine’s behaviour. A richer action space can allow the machine to explain its reasoning by presenting its predictions to the human. An adaptive human can also learn a better $O_h$ through interaction, which opens further theoretical questions such as how the learning rate of the human and the machine affect the belief contraction. In the most general case, the human would use a BRM and model the machine just as the machine models them, which leads to an infinite recursion of beliefs. Previous work indicates that humans on average do not recurse too deep when reasoning about others \cite{Doshi2010ModelingRR}, and a fixed-depth recursion is still tractable within our framework thanks to \citet{Zettlemoyer2008MultiAgentFW}.

%%%%%%%%%%%%%%%%%%%%%%%%%%%%%%%%%%%%%%%%%%%%%%%%%%%%%%%%%%%%%%%%%%%%%%%%

%%%%%%%%%%%%%%%%%%%%%%%%%%%%%%%%%%%%%%%%%%%%%%%%%%%%%%%%%%%%%%%%%%%%%%%%

%%% The acknowledgments section is defined using the "acks" environment
%%% (rather than an unnumbered section). The use of this environment 
%%% ensures the proper identification of the section in the article 
%%% metadata as well as the consistent spelling of the heading.
\begin{acks}
This work was supported by: the Academy of Finland (Flagship programme:  Finnish Center for Artificial Intelligence; decision 828400),\begin{wrapfigure}{r}{0.15\columnwidth}
\begin{center}
\vspace{-5pt}
    \includegraphics[width=0.15\columnwidth]{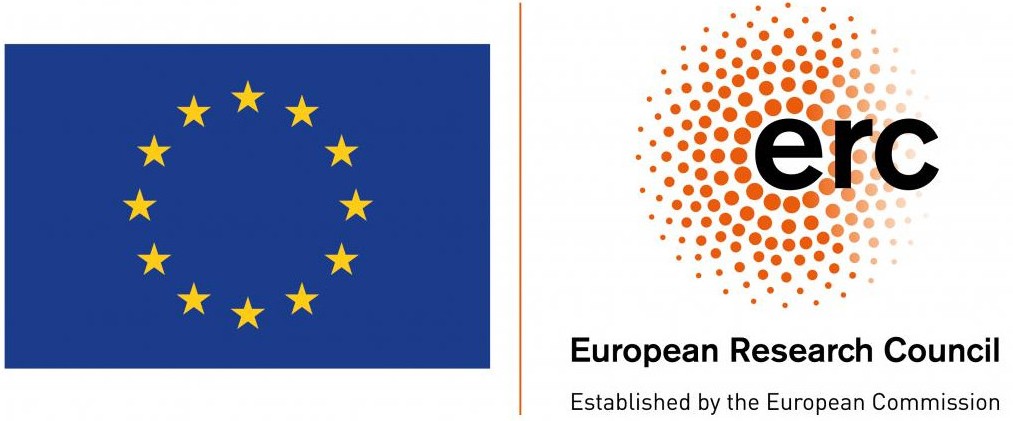}
\end{center}
\end{wrapfigure}
the European Research Council (ERC) under the European Union's Horizon 2020 research and innovation programme (grant agreement No.~758824 \textemdash INFLUENCE), 
the UKRI Turing AI World-Leading Researcher Fellowship EP/W002973/1, ELISE travel grant (GA no 951847), KAUTE Foundation, and the Aalto Science-IT Project.

%\begin{center}
%\vspace{-5pt}
%    \includegraphics[width=0.17\columnwidth]{LOGO_ERC-FLAG_EU_.jpg}
%\end{center}

\end{acks}

%%%%%%%%%%%%%%%%%%%%%%%%%%%%%%%%%%%%%%%%%%%%%%%%%%%%%%%%%%%%%%%%%%%%%%%%

%%% The next two lines define, first, the bibliography style to be 
%%% applied, and, second, the bibliography file to be used.

\bibliographystyle{ACM-Reference-Format} 
\bibliography{ms}

%%%%%%%%%%%%%%%%%%%%%%%%%%%%%%%%%%%%%%%%%%%%%%%%%%%%%%%%%%%%%%%%%%%%%%%%

\end{document}

\maketitle

\section{Proofs}
\subsection{The Machine-optimistic Human from Decision-theoretic Principles}
\paragraph{Human's Best-response Model.} 
For a given model space of the machine, $\mathcal{M}_m$, the best-response model of the human is defined as \linebreak $BRM_h(STM_h, \mathcal{M}_m) = ( \Bar{\mathcal{S}}_h, \Bar{\mathcal{A}_h}, \Bar{D}_h, \Bar{\Omega}_h, \Bar{R}_h)$ where $\Bar{\mathcal{S}}_h = \mathcal{S} \times \mathcal{A}_m \times \mathcal{I}_m$, $\Bar{\Omega}_h = \Omega_h \times \mathcal{A}_m$, and $\Bar{\mathcal{A}}_h = \mathcal{A}_h \cup \{\textit{noop}\}$. The state and observation spaces are augmented with $\mathcal{A}_m$ to allow the human to observe the machine's actions. The augmented reward function is defined as $\Bar{R}_h(\Bar{s}_h, \Bar{a}_h) = R_h(s, a_c(a_m, \Bar{a}_h)) - \mathbb{I}[\Bar{a}_h \neq \textit{noop}]c_h(s, \Bar{a}_h)$. In most cases, the natural way to define the $c_h$ is as a constant $c$. However, this can be relaxed as long as the function $c_h$ satisfies the condition $arg\max_{a \in \mathcal{A}_h}Q^{\pi^*_h}(b,a) = arg\max_{a \in \mathcal{A}_h}Q^{\pi^*_h}(b,a) - c_h(b,a)$ for all $b \in \Delta(\mathcal{S})$, where $Q^{\pi^*_h}(b,a)$ is the optimal value function of $STM_h$. The dynamics of $BRM_h$ is defined as follows:
\begin{align}
    \Bar{D}_h(\Bar{s}'_h, \Bar{o}_h \mid \Bar{s}_h, a_h) = \sum_{\Bar{o}_m}  &\Bar{O}(\Bar{o}_h, \Bar{o}_m \mid \Bar{s}'_h, a_c(a_m, \Bar{a}_h))T_h(s' \mid s, a_c(a_m, \Bar{a}_h)) \nonumber \\ 
    &\underbrace{\beta_m(I'_m \mid I_m, a_c(a_m, \Bar{a}_h), \Bar{o}_m)\Bar{\pi}_m(a'_m \mid I'_m)}_\text{Machine Term}.
    \label{eq:humanbrm}
\end{align}

The important difference between the dynamics defined by the equation \eqref{eq:humanbrm} and the dynamics of the machine's BRM is what the policy of the other agent is used for. Since the human already observes $a_m$ before deciding on $\Bar{a}_c$, they do not need $\Bar{\pi}_m$ to predict that. Instead, here $\beta_m$ and $\Bar{\pi}_m$ give the probability of machine's next action $a'_m$ which contributes to the transition probability since the $a'_m$ is part of the $\Bar{s}'_h$. 

The machine-optimistic human believes that the machine's behaviour will align perfectly with them from $t+1$ onwards, which means the $\Bar{\pi}_m(a'_m \mid I'_m)$ will execute the same actions with the human's optimal policy for their subjective task model, $\pi^*_h$. In that case, clearly the term marked as \emph{Machine Term} in equation \eqref{eq:humanbrm} can be entirely replaced by $\pi^*_h$. This also eliminates the need for modelling $\Bar{o}_m$, since it no more contributes to the probabilities.

Let the current state at $t$ be $\Bar{s}^{(t)}_h$, where the machine's action is $a^{(t)}_m$ and the human's belief in their subjective POMDP $M_h$ be $b^{(t)}_h \in  \Delta(\mathcal{S})$. Important to remember that the augmented state-space is $\Bar{\mathcal{S}}_h = \mathcal{S} \times \mathcal{A}_m \times \mathcal{I}_m$, and the $\mathcal{A}_m$ part is fully-observed. Since the machine-optimistic human does not keep track of $\mathcal{I}_m$ due to its approximation of the future, $b^{(t)}_h$ is the sufficient belief in the BRM.

If the machine behaves just as the $\pi^*_h$ would have from $t+1$ on, we can write the Q-values as:

\begin{align}
    Q(b^{(t)}_h, \Bar{a}_h) &= \mathbb{E}[R_h(s, a_c(a_m, \Bar{a}_h)) - \mathbb{I}[\Bar{a}_h \neq \textit{noop}]c_h(s, \Bar{a}_h) + V^{\pi^*_h}(b^{(t+1)}_h)] \\
    &=\mathbb{E}[R_h(s, a_c(a_m, \Bar{a}_h)) + V^{\pi^*_h}(b^{(t+1)}_h)] - \mathbb{I}[\Bar{a}_h \neq \textit{noop}]\mathbb{E}[c_h(s, \Bar{a}_h)]
    \label{eq:onestep}
\end{align}
where $V^{\pi^*_h}(b^{(t+1)}_h)$ is the optimal value function of the $M_h$ computed according to $OPT_h$, and the expectation is over both $b^{(t)}_h$ and the dynamics. In other words, $Q(b^{(t)}_h, \Bar{a}_h)$ is the one step look-ahead action values where the return from $t+1$ onwards is estimated with $V^{\pi^*_h}(b^{(t+1)}_h)$.

If \textit{noop} is played, there is no cost of effort, $a_m$ gets executed and then the continuation happens according to $\pi^*_h$. Therefore, it follows from equation \eqref{eq:onestep} that:
 \begin{align}
      Q(b^{(t)}_h, \textit{noop}) = \mathbb{E}[R_h(s, a_m) + V^{\pi^*_h}(b^{(t+1)}_h)] = Q^{\pi^*_h}(b^{(t)}_h, a_m).
 \end{align}
Let us assume \textit{noop} is not the optimal $\Bar{a}_h$, and denote the optimal action here with $\Bar{a}^*_h$ where $\Bar{a}^*_h \neq \textit{noop}$. Then the inequality $\mathbb{E}[R_h(s, \Bar{a}^*_h) + V^{\pi^*_h}(b^{(t+1)}_h)] - \mathbb{E}[c_h(s, \Bar{a}^*_h)] > Q^{\pi^*_h}(b^{(t)}_h, a_m)$ must hold, which is equivalent to:
\begin{align}
    Q^{\pi^*_h}(b^{(t)}_h, \Bar{a}^*_h) - \mathbb{E}[c_h(s, \Bar{a}^*_h)] > Q^{\pi^*_h}(b^{(t)}_h, a_m).
    \label{eq:inequality_1}
\end{align}
In subsection 3.1, we have stated that any assumption on $c_h$ can be relaxed as long as it satisfies the condition $arg\max_{a \in \mathcal{A}_h}Q^{\pi^*_h}(b,a) = arg\max_{a \in \mathcal{A}_h}Q^{\pi^*_h}(b,a) - c_h(b,a)$ for all $b \in \Delta(\mathcal{S})$ where $c_h(b,a) = \mathbb{E}_{s \sim b}[c_h(s,a)]$. It follows from this condition that $arg\max_{\Bar{a}_h}{Q^{\pi^*_h}(b^{(t)}_h, \Bar{a}_h) - \mathbb{E}[c_h(s, \Bar{a}_h)]} = \pi_h^*(b^{(t)}_h)$. Therefore, $Q^{\pi^*_h}(b^{(t)}_h, \Bar{a}^*_h) = Q^{\pi^*_h}(b^{(t)}_h, \pi_h^*(b^{(t)}_h)) = V^{\pi^*_h}(b^{(t)}_h)$. This implies the condition for overriding given in equality \eqref{eq:inequality_1} is equivalent to:
\begin{align}
     V^{\pi^*_h}(b^{(t)}_h) - Q^{\pi^*_h}(b^{(t)}_h, a_m) >  \mathbb{E}[c_h(s, \Bar{a}^*_h)]
    \label{eq:inequality_final}
\end{align}
and the $\Bar{a}^*_h =  \pi_h^*(b^{(t)}_h) \blacksquare$

\subsection{Proofs for Section 6}

\begin{customlm}{6.3}
Let $\mathbf{T}^a$ be the deterministic transition matrix with entries $(i,j)$ as $T(s_i \mid s_j, a)$ where $\mathbb{T}^a(b) = \mathbf{T}^ab$. If $rank(\mathbf{T}^a) = \mid S \mid$ for all $a \in \mathcal{A}$, then $KL(\mathbb{T}^a(b_m) \mid \mid \mathbb{T}^a(b_h)) = KL(b_m \mid \mid b_h)$ for all $a \in \mathcal{A}$ and $b_h, b_m \in \Delta(S)$. Therefore, $rank(\mathbf{T}^a) < \mid S \mid$ is a necessary condition for the strict inequality.
\end{customlm}

\paragraph{Proof of lemma 6.3} A deterministic transition matrix $\mathbf{T}^a$ is an $\mid S \mid \times \mid S \mid$ matrix where each row has a single non-zero entry, which equals $1$. If $rank(\mathbf{T}^a) = \mid S \mid$, then this implies the same holds for columns, that each column has a single non-zero entry, which equals $1$. Therefore, $\mathbf{T}^a$ is a \emph{permutation matrix}. The multiplication $\mathbf{T}^a b$ simply permutes the columns of $b$ in a specific order. Since the two vectors $b_1$ and $b_2$ are permuted in the same order, $KL(\mathbf{T}^a b_1 \mid\mid \mathbf{T}^a b_2) = KL(b_1 \mid\mid b_2) \blacksquare$. 

\begin{customthm}{6.4}
For  $STM_h = (\mathcal{S}, \mathcal{A},T, \Omega, R, O_h, OPT)$ and \sloppy $STM_m = (\mathcal{S}, \mathcal{A},T, \Omega, R, O_m, OPT)$ with $O_m = O$, let $KL(O_m(. \mid s,a) \mid \mid O_h(. \mid s,a)) \leq \epsilon_O$ for all $(s,a) \in \mathcal{S} \times \mathcal{A}$. Let  $\mathbf{T}^{a_c}$ be the transition matrix of the hidden Markov model induced by a fixed centaur policy $\pi_c$, with entries $T(s' \mid s, a_c)$ where $a_c$ denotes the executed centaur action. The belief updates satisfy the inequality; 

     \begin{align*}
         \mathbb{E}_{o \sim O_m(. \mid b_m, a_c)}[KL(&b'_m \mid \mid b'_h)] \leq
         \left( 1 - \alpha(\mathbf{T}^{a_c}) \right)KL(b_m \mid\mid b_h) \\ &+\gamma(O_m)3\sqrt{\epsilon_O} -  \left( \frac{\gamma(O_m)KL(b_m \mid \mid b_h)}{\sqrt{2}\log{\frac{1}{\mu}}}\right)^2,
     \end{align*} where $\gamma(O_m)> 0$ is the induced HMM's value of observation, the $\alpha(\mathbf{T}^{a_c})$ is its minimal mixing rate, and $\mu$ is a constant such that $b_h(s), b_m(s) \geq \mu$ for all $s$.

\end{customthm} 

\paragraph{Proof of Theorem 6.4} We need the following results from previous work for our proof. 
\subparagraph{Theorem 3 of \citet{Boyen1998TractableIF}} Let $\mathbf{T}^a$ be a stochastic transition matrix with entries $(i,j)$ as $T(s_i \mid s_j, a)$, the $b_1$ and $b_2$ be two distributions over the state space $\mathcal{S}$, and $\alpha(\mathbf{T}^a)$ be the minimal mixing rate of $\mathbf{T}^a$. Then 
$KL(\mathbf{T}^a b_1 \mid\mid \mathbf{T}^a b_2) \leq \left( 1 - \alpha(\mathbf{T}^a) \right)\left( KL(b_1 \mid\mid b_2) \right)$.
\subparagraph{Lemma 3.9 of \citet{EvenDar2007TheVO}} For a hidden Markov model, let $O_1$ be the true observation model  and $O_2$ be an approximation with $KL(O_1(.\mid s) \mid\mid O_2(.\mid s)) \leq \epsilon_{O}$ for all $s \in \mathcal{S}$. Let $(\mathbb{O}^{o}_ib)(s) \triangleq \frac{b(s)O_i(o \mid s)}{\sum_{s'}b(s')O_i(o \mid s')}$ and  $O_i(.\mid b) \triangleq \sum_{s'}b(s')O_i(o \mid s')$. For every pair of belief states $b_1$ and $b_2$,
\begin{align*}
    \mathbb{E}_{o \sim O_1(. \mid b_1)} &\left [KL(\mathbb{O}^{o}_1 b_1 \mid\mid \mathbb{O}^{o}_2 b_2)\right] \leq \\
    & KL(b_1 \mid\mid b_2) + \epsilon_O - KL(O_1(.\mid b_1) \mid\mid O_2(.\mid b_2)).
\end{align*}

\subparagraph{Proposition 3.5 of \citet{EvenDar2007TheVO}} Let $\gamma(O)$ be the value of observation of a hidden Markov model with observation model $O$, $\hat{O}$ be an approximate observation model such that $KL(O(. \mid s) \mid\mid \hat{O}(. \mid s)) \leq \epsilon_O$ for all $s$, $b_1$ and $b_2$ be belief states such that $b_1(s), b_2(s) \geq \mu$ for all $s$ for a $\mu > 0$. Then,
\begin{align*}
    KL&(O(. \mid b_1) \mid\mid \hat{O}(. \mid b_2)) \geq \\ & \frac{1}{2} \left(\frac{\gamma(O) KL(b_1 \mid\mid b_2)}{-\log\mu}\right)^2 +\epsilon_O - \gamma(O)3\sqrt{\epsilon_O}.
\end{align*}

\paragraph{The construction of the proof for Theorem 6.4.} In theorem 6.4, for simplicity we have assumed that $O_m = O$, so the machine's observation model is correct. Note that the full belief update of an agent is written as $b'= \mathbb{O}^{o,a}_i(\mathbb{T}^a_i(b))$ with the operators defined as $(\mathbb{T}^a_ib)(s) = \sum_{s'}{b(s')T_i(s \mid s',a)}$ and $(\mathbb{O}^{o,a}_ib)(s) = \frac{b(s)O_i(o \mid s)}{\sum_{s'}b(s')O_i(o \mid s')}$. Thus we can write the belief differences after an update with action-observation $(a_c, o)$ as 
\begin{align}
    KL(\mathbb{O}^{o,a_c}_m(\mathbb{T}^{a_c}(b_m)) \mid\mid \mathbb{O}^{o,a_c}_h(\mathbb{T}^{a_c}(b_h))),
    \label{eqn:beliefupdate}
\end{align}
where the $a_c$ is fixed due to the fixed centaur policy $\pi_c$. Applying the lemma 3.9 of \citet{EvenDar2007TheVO} to the equation \ref{eqn:beliefupdate} gives:

\begin{align}
      \mathbb{E}_{o \sim O_m(. \mid b_m)}&\left[KL(\mathbb{O}^{o,a_c}_m(\mathbb{T}^{a_c}(b_m)) \mid\mid \mathbb{O}^{o,a_c}_h(\mathbb{T}^{a_c}(b_h)))\right]  \leq \nonumber \\
      & KL(\mathbb{T}^{a_c}(b_m) \mid\mid \mathbb{T}^{a_c}(b_h)) + \epsilon_O - KL(O_m(. \mid b_m) \mid\mid O_h(. \mid b_h)).
\end{align}

Applying the proposition 3.5 of \citet{EvenDar2007TheVO} to the equation 8's right-most term gives:

\begin{align}
      \mathbb{E}_{o \sim O_m(. \mid b_m)}&\left[KL(\mathbb{O}^{o,a_c}_m(\mathbb{T}^{a_c}(b_m)) \mid\mid \mathbb{O}^{o,a_c}_h(\mathbb{T}^{a_c}(b_h)))\right]  \leq \nonumber \\
      & KL(\mathbb{T}^{a_c}(b_m) \mid\mid \mathbb{T}^{a_c}(b_h)) - \frac{1}{2} \left(\frac{\gamma(O_m) KL(b_m \mid\mid b_h)}{-\log\mu}\right)^2 + \gamma(O_m)3\sqrt{\epsilon_O}.
\end{align}

Applying the theorem 3 of \citet{Boyen1998TractableIF} to the term $KL(\mathbb{T}^{a_c}(b_m) \mid\mid \mathbb{T}^{a_c}(b_h))$ in equation 9 gives:

\begin{align}
      \mathbb{E}_{o \sim O_m(. \mid b_m)}&\left[KL(\mathbb{O}^{o,a_c}_m(\mathbb{T}^{a_c}(b_m)) \mid\mid \mathbb{O}^{o,a_c}_h(\mathbb{T}^{a_c}(b_h)))\right]  \leq \nonumber \\
      &  \left( 1 - \alpha(\mathbf{T}^{a_c}) \right)\left( KL(b_m \mid\mid b_h) \right)
     - \frac{1}{2} \left(\frac{\gamma(O_m) KL(b_m \mid\mid b_h)}{-\log\mu}\right)^2 + \nonumber \\ &\gamma(O_m)3\sqrt{\epsilon_O}.
\end{align}

Simplifying the operator notation for the updated beliefs, we get:

\begin{align}
      \mathbb{E}_{o \sim O_m(. \mid b_m)}&\left[KL(b'_m \mid\mid b'_h)\right]  \leq \nonumber \\
      &  \left( 1 - \alpha(\mathbf{T}^{a_c}) \right)\left( KL(b_m \mid\mid b_h) \right)
     - \frac{1}{2}\left(\frac{\gamma(O_m) KL(b_m \mid\mid b_h)}{-\log\mu}\right)^2 + \nonumber \\
     &\gamma(O_m)3\sqrt{\epsilon_O} \qquad \blacksquare
\end{align}

\section{Experimental Details}
\paragraph{Food Truck Environment.} The state space $\mathcal{S}$ consists of regular grids and restaurant grids. Each restaurant grid consists of a chain of three states: the entry, the exit, and the terminal state. For instance, when the agent visits the vegan restaurant grid it first transitions to \emph{vegan-entry} state. Then at next time-step, regardless of the taken action, it transitions to the \emph{vegan-exit} state, and then at next time-step it automatically transitions to the absorbing terminal state. The rewards are: vegan-entry: $-10$, vegan-exit: $+20$, doughnut-entry: $+10$, doughnut-exit: $-10$. Every time-step has $-0.1$ cost. Thus, if an agent enters a doughnut shop, it first receives $+10 - 0.1$ reward, then after a time-step receives $-10 - 0.1$, and then transitions to the terminal state.

The belief is discretized with $2000$ particles. We created 200 $\gamma$s from the low $\gamma$ region and 200 from the high $\gamma$ region using a grid. Then applied cross-product with five different $c_h$ values representing low, medium, and high. The belief converges to the true human MDP, however the MCTS in this case needs a lot of iterations to converge to the optimal policy with random rollout, considering the environment needs long exploration and is deterministic. Since the machine already knows its subjective goal (to reach the vegan restaurant), we used the time-step distance to vegan restaurant as a heuristic to evaluate the states. This does not use any information that the machine does not know.

\paragraph{Food Shelter Environment.} The shelter collapse probability is $0.1$. Food reward is $1.0$, shelter collapse cost is $-0.1$, step cost is $0.0$. The agent starts at the shelter, and the first food appears at $(1,1)$. The beliefs are discretized with $1200$ particles, where there are 12 $\epsilon$s generated as a grid in $[0.0, 0.45]$ and 100 $c_h$s in $[0.0, 0.5]$. Machine's Q function is used as the MCTS evaluation heuristic as a more informative alternative to random rollouts.

\paragraph{Computing and software resources} All experiments were run on a laptop with the following specs: 
\begin{itemize}
    \item Processor: Intel(R) Core(TM) i7-8550U CPU @ 1.80GHz 2.00 GHz
    \item Installed RAM: 16,0 GB (15,7 GB usable).
\end{itemize}

Overall, both experiments take about 12 hours to run on a single core. The JuliaPOMDPs libraries have been used to implement and run the experiments.

\section{Discussion on Surrogate Models and Behavioural Equivalence Classes}
In both of our experiments, the difference between the subjective task model of the human $STM_h$ and the machine $STM_m$ was represented by a parameter. In Food Truck, this is the discount parameter $\gamma$, whereas it is action noise $\epsilon$ in Food Shelter. The general idea here is that we assume there is a structure in the space of human's subjective task models, and $STM_h$ can be determined by a vector of hidden parameters $\theta$. Since the machine-optimistic human uses their optimal policy for the $STM_h$, $\pi^*_h$, to override the machine, we can use the structure in $\theta$ to predict the human's overriding policy $\Bar{\pi}_h$.   

For instance, in the Food Shelter experiment we can generate $1000$ STMs by sampling $(\epsilon^{(i)}, c^{(i)}_h), i \in 1,...,1000$ pairs from a prior. Then, we solve each and generate $\pi^{(i)*}_h$. After this point, we choose a machine policy $\pi_m$ such that the $\pi_m$ tries each action at each state with non-zero probability. We pair each $\pi^{(i)*}_h$ with $\pi_m$ to generate the trajectories from $BRM_m$ for each $(\epsilon^{(i)}, c^{(i)}_h)$ pair. Then a neural network can be fitted to this dataset so that given $(\epsilon, c_h, s)$ or $(\epsilon, c_h, b_h)$ it predicts $\Bar{a}_h$. There are finitely many behavioural equivalence classes for an optimist, since the space of all possible $(\epsilon, c_h)$ pairs maps to a finite and discrete set of optimal policies $\Pi^*_h$. Therefore, intuitively the neural network trained as above would need to learn to classify an $(\epsilon, c_h)$ to the correct equivalence class, and then imitate the policy of the class.

Once the neural network (NN) is trained, the weights of newly added particles can be computed with the NN, and if these particles are sampled for a simulation, the action predictions of the NN can be used to approximate $\Bar{\pi}_h$. In this case, the accuracy of both the belief approximation and the simulations will depend on how good the neural network can approximate $\Bar{\pi}_h$ for new particles and how often the particle reinvigoration will be necessary.
\section{Further Experiments}
\begin{figure}
  \begin{center}
    \includegraphics[width=0.5\textwidth]{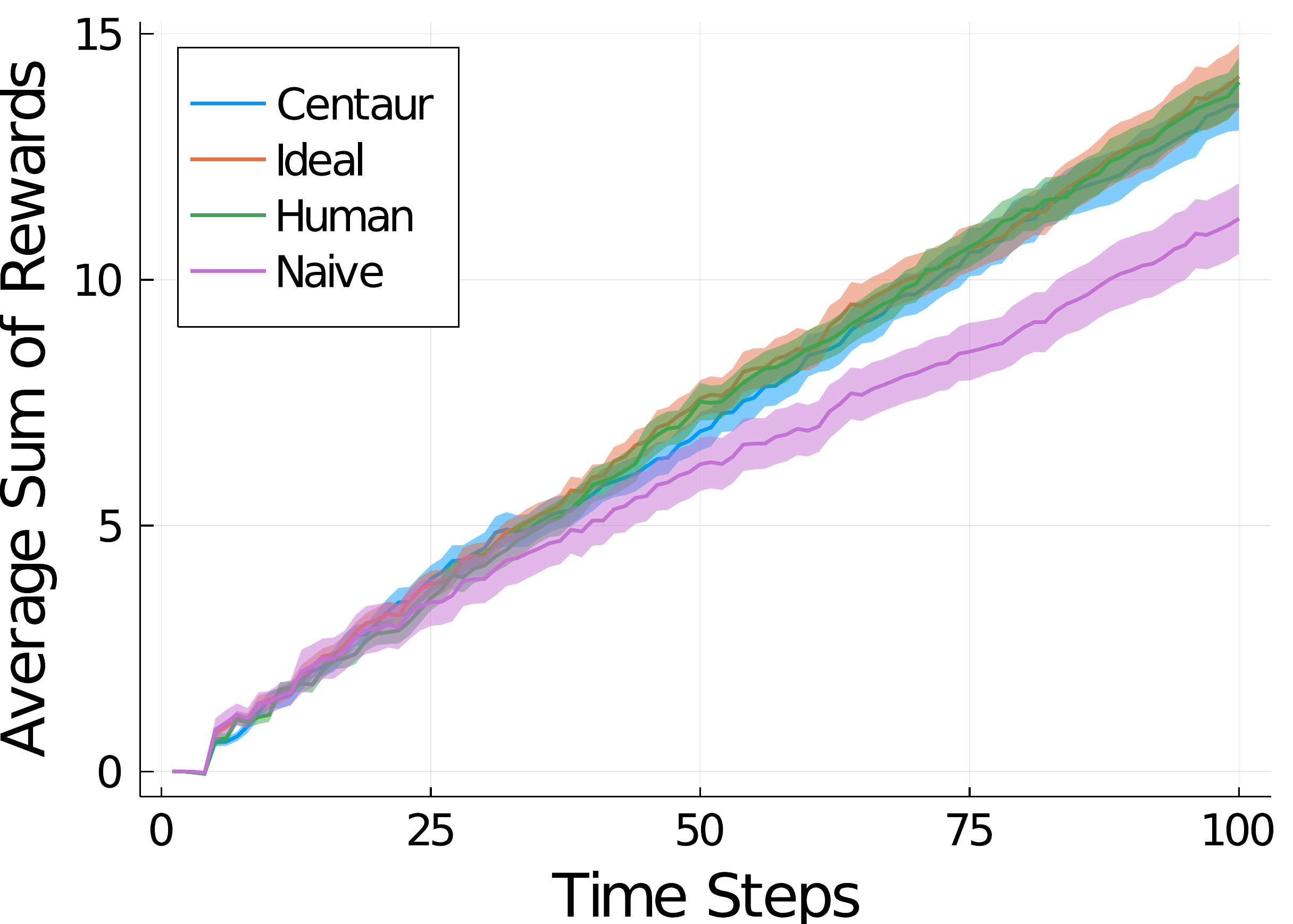}
  \end{center}
  \caption{Food Shelter experiment when both the human and the machine have the correct model.}
  \label{fig:bothcorrect}
\end{figure}

\paragraph{When both agents are correct.} When both the human and the machine have the correct model,  $STM_h=STM_m=OTM$, their models fully agree. In this case, we expect the \emph{centaur} to be able to eventually learn that the human's model agrees with the machine, and perform the same as the human's policy. We have tested this case in the Food Shelter environment where $\epsilon=0$, so the human does not demonstrate maladaptive avoidance and has the correct transition model. Everything else are kept the same as in the main experiment. In figure \ref{fig:bothcorrect}, we see that indeed the centaur
 ideal, and human perform very similarly. Perhaps more surprisingly, when the machine ignores the human and acts as if it is alone (naive), it performs worse. This is because, even though the $STM_h = STM_m$, the machine is using an online MCTS-variant to choose its next action. With the number of iterations we have run, MCTS does not provide the exact optimal policy. We could have increased the number of iterations, however we believe this is an interesting result showing even with both models correct, the centaur can still benefit from human's help when its planning algorithm is not exact.
 
\paragraph{When the machine is wrong, but the human is correct.} This is the case when $STM_h=OTM$, but $STM_m \neq OTM$. We consider this with the Food Shelter environment by simply swapping the models of the human and the machine. So, the human has the correct $T$ with action noises $0.1$, whereas the machine over-estimates these as $\epsilon + 0.1$ for vertical and horizontal moves, and $2\epsilon + 0.1$ for diagonal moves. Everything else is identical with the experiment in the manuscript. In this case, \emph{naive} performs very badly since the machine's STM is wrong, leading to a bad policy, which is overridden by the human a lot. Machine ends up paying a lot of cost for getting overridden ($c_m=0.2$).  
\begin{figure}
  \begin{center}
    \includegraphics[width=0.5\textwidth]{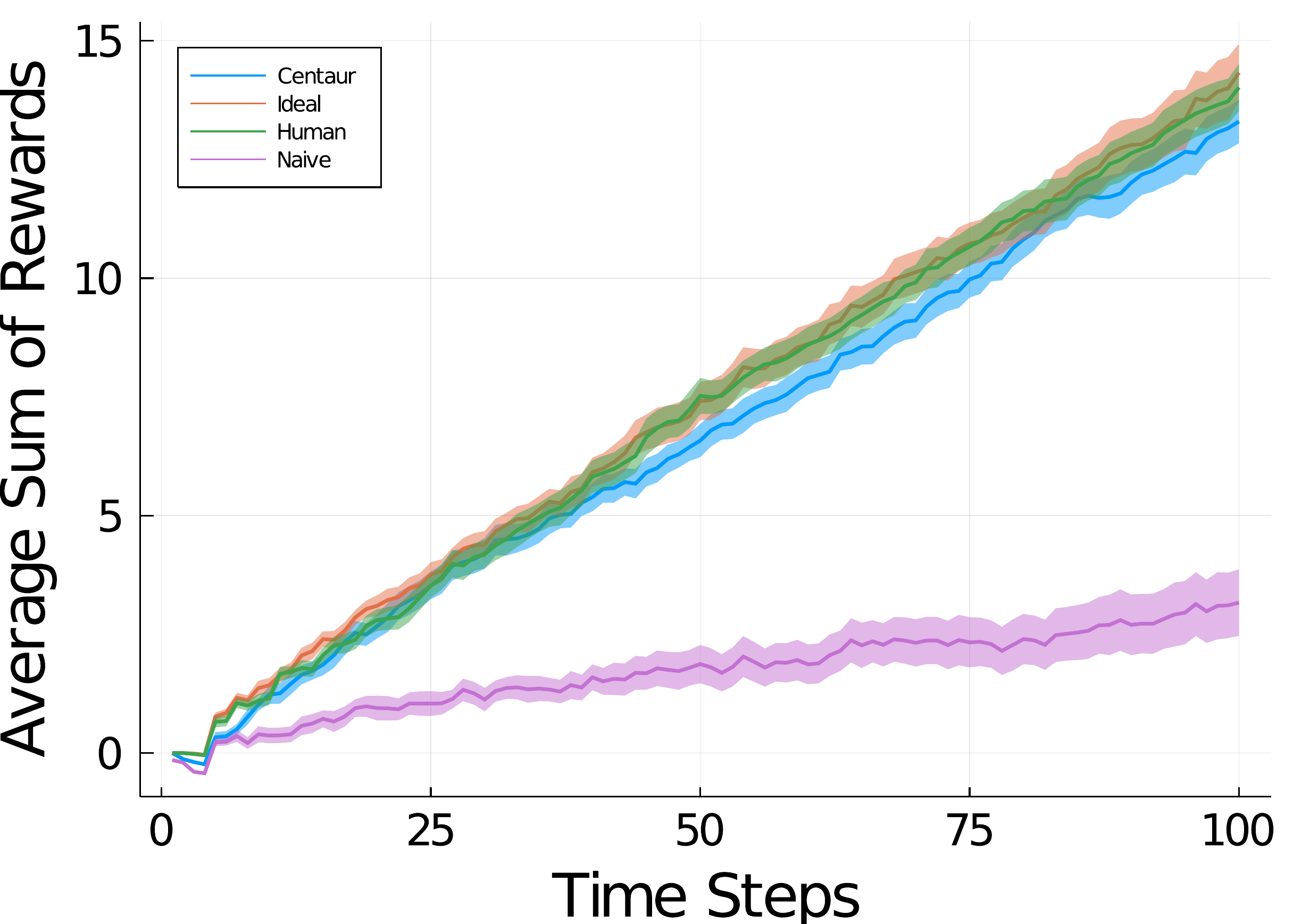}
  \end{center}
  \caption{Food Shelter experiment when the human has the correct model, but the machine has the wrong one.}
  \label{fig:machinewrong_humanright}
\end{figure}
The \emph{human} performs the best since they have the correct model. The most interesting cases are the \emph{ideal} and the \emph{centaur}. In ideal, the machine knows the model of the human, but still wrong about the model of the task. However, since it knows when will human override it, the machine is able to learn a policy that avoids getting overridden by the human, which performs the same as the human's policy. The centaur does not know the model of the human, and infers it from interaction. Clearly, it can do this quickly and perform similar to the ideal. This shows that if the machine is wrong but the human is right, the human can correct the machine's behaviour with overrides and help it learn a better policy.

\bibliographystyle{unsrtnat}
\bibliography{ms}

%Optionally include extra information (complete proofs, additional experiments and plots) in the appendix.
%This section will often be part of the supplemental material.